\documentclass{article}
\usepackage{amsthm}

    \PassOptionsToPackage{numbers, compress}{natbib}

 \usepackage[preprint]{neurips_2026}

\usepackage[utf8]{inputenc} %
\usepackage[T1]{fontenc}    %
\usepackage{hyperref}       %
\usepackage{url}            %
\usepackage{booktabs}       %
\usepackage{amsfonts}       %
\usepackage{nicefrac}       %
\usepackage{microtype}      %
\usepackage{xcolor}         %
\usepackage{colortbl}       %

\usepackage{wrapfig}

\usepackage{graphicx}

\newcommand{\sS}{\mathcal{S}}
\newcommand{\sA}{\mathcal{A}}
\newcommand{\sP}{\mathcal{P}}
\newcommand{\sR}{\mathcal{R}}

\newcommand{\sT}{\mathcal{T}}
\newcommand{\sW}{\mathcal{W}}

\definecolor{flowcolor}{RGB}{0, 102, 204} %
\newcommand{\flow}[1]{\textbf{\textcolor{flowcolor}{\texttt{#1}}}}
\usepackage{subcaption}
\usepackage{multirow}
\usepackage{bm}
\usepackage{dsfont}
\usepackage{amsmath}
\usepackage{amssymb}
 \definecolor{mplblue}{HTML}{1F77B4}

\title{DiPRL: Learning Discrete Programmatic Policies via \\Architecture Entropy Regularization}

\author{%
  Chengpeng Hu$^{1}$ \quad Yingqian Zhang$^{1}$ \quad Hendrik Baier$^{1,2}$\\
  $^{1}$Eindhoven University of Technology, Eindhoven, the Netherlands\\
  $^{2}$Centrum Wiskunde \& Informatica, Amsterdam, the Netherlands
}

\begin{document}

\maketitle

\begin{abstract}
Programmatic reinforcement learning (PRL) offers an interpretable alternative to deep reinforcement learning by representing policies as human-readable and -editable programs. While gradient-based methods have been developed to optimize continuous relaxations of programs, they face a significant performance drop when converting the continuous relaxations back into discrete programs. Post-hoc discretization can discard optimized branches and parameters in a program, which results in a collapse of policy expressivity and lowered task performance, leading in turn to a need for additional fine-tuning. To overcome these limitations, we propose Differentiable Discrete Programmatic Reinforcement Learning (DiPRL), a method that learns programmatic policies that become nearly discrete during training, avoiding a separate post-hoc fine-tuning stage. We first analyze the inherent risks of performance drop introduced by post-hoc discretization of gradient-based methods. Then, we introduce programmatic architecture entropy regularization, which enables smooth, differentiable training that encourages convergence toward a discrete program. DiPRL maintains the efficiency of gradient-based optimization while mitigating the risks of post-hoc discretization. Our experiments across multiple discrete and continuous RL tasks demonstrate that DiPRL can achieve strong performance via interpretable programmatic policies.
\end{abstract}

\section{Introduction}

Programmatic reinforcement learning (PRL) represents policies as human-readable programs, using explicit control flow (e.g., if-else) to choose primitive actions (e.g., ``left'' and ``right'' in CartPole-v1). PRL has achieved performance comparable to the neural policies produced by deep reinforcement learning~\cite{mnih2015human} across domains such as games~\cite{verma2019imitation,hasanbeig2021deepsynth,aleixo2023show,marino2021programmatic,cao2022galois} and continuous control~\cite{verma2018programmatically,inala2020synthesizing,qiu2022programmatic}, while offering interpretability. Fig.~\ref{fig:program_demo1} illustrates an example of a simple program to control CartPole-v1 with only one ``if-else'' statement.

A common approach to learning programmatic policies is local search (LS), which initializes a random program and iteratively modifies it using the production rules of a language grammar~\cite{marino2021programmatic}. However, LS is inefficient because transitions collected during program evaluations are discarded. LS also requires additional parameter optimizers such as Bayesian Optimization (BO) for domains like continuous control, where program conditions defined on state features  have parameters that must be learned~\cite{verma2018programmatically}. To alleviate these problems, $\pi$-PRL~\cite{qiu2022programmatic} relaxes a discrete program into a continuous, differentiable derivation tree, which allows the use of collected transitions for policy gradient training of both program architecture and parameters. This tree assigns probabilities to program branches and actions, and aggregates all possible execution paths to compute an action distribution. After policy gradient training, a discrete resulting program architecture has to be sampled from the tree via maximum likelihood, which then requires further fine-tuning.

However, a significant issue of the differentiable derivation tree is the risk of performance drop during post-hoc discretization. Converting a continuous derivation tree into a discrete program inevitably discards some program branches, which can potentially result in the loss of important parameters and branches that contribute to the performance of the policy. In extreme cases, the derivation tree can collapse into a trivial program architecture that severely limits its expressivity. In this case, even additional fine-tuning may not improve the extracted program any further. We present this phenomenon as a motivating example for our work in Section~\ref{sec:motivation}, where a program with fine-tuning never recovers its performance after the discretization in the continuous task, Ant RandomGoal.

Our main contributions are as follows:

(1) We show that post-hoc discretization of continuous derivation trees can cause significant performance drops by discarding useful program branches, and that additional fine-tuning does not always recover the lost performance.

(2) We propose \textbf{Differentiable Discrete Programmatic Reinforcement Learning (DiPRL)}, which augments the standard differentiable derivation trees via \emph{program architecture regularization}. DiPRL enables end-to-end learning of programs by encouraging convergence to discrete architectures during training, which eliminates the need for a separate post-discretization fine-tuning stage.

(3) We conduct experiments on discrete and continuous RL tasks to compare DiPRL with baselines. The results show that DiPRL gradually converges to discrete programmatic policies throughout training, allowing it to achieve performance comparable to neural policies while maintaining the interpretability inherent in programmatic representations.

\section{Related Work}

\begin{figure}[b]
    \centering
    \scalebox{0.9}{
        \fbox{
            \parbox{0.8\linewidth}{
                \begin{align*}
                &\flow{if}~\left( -0.16-0.31 \cdot \bm{f_0} - 2.1\cdot \bm{f_1} - 6.6\cdot\bm{f_2}-2.3\cdot\bm{f_3} > 0 \right): \\
                & \qquad \text{Left} \\
                &\flow{else}\\
                & \qquad \text{Right}
                \end{align*}
            }
        }
    }
    \caption{A programmatic policy discovered by DiPRL for CartPole-v1. $f_0, f_1, f_2, f_3$ denote symbolic features corresponding to cart position, cart velocity, pole angle, and pole angular velocity, respectively. The program splits the state space into two regions and selects either the ``left'' or ``right'' action. It achieves the maximum score of 500 on CartPole-v1.}
    \label{fig:program_demo1}
\end{figure}

Programmatic reinforcement learning encodes policies as programs~\cite{choi2005learning,acharya2023neurosymbolic}. Fig.~\ref{fig:program_demo1} presents an example of a programmatic policy for CartPole-v1. While often providing competitive performance to neural policies~\cite{qiu2022programmatic}, PRL enables humans to interpret the learned policies, edit them, and verify their correctness~\cite{wang2023verification}. Its performance has been demonstrated in various domains such as 
games~\cite{verma2019imitation,hasanbeig2021deepsynth,aleixo2023show,marino2021programmatic,cao2022galois,bashir2023assessing,moraes2023choosing,moraes2024searching}, Karel tasks~\cite{trivedi2021learning,liu2023hierarchical,liu2025synthesizing,lin2024hierarchical}, robotics~\cite{verma2018programmatically,inala2020synthesizing,qiu2022programmatic,Wu2024Synthesizing}, and real-world optimization problems~\cite{gu2024pi,gu2025procc}. Decision tree (DT) policies can also be expressed as programs, for example by converting a tree into equivalent \texttt{if-else} statements. However, DTs are most commonly learned in supervised settings~\cite{verwer2019learning,vos2021efficient,WAN2021nbdt}. In RL, most DT approaches rely on imitation learning~\cite{verma2018programmatically,silver2020few}, which can suffer from distillation gaps~\cite{qiu2022programmatic}. Recent differential DTs such as~\cite{silva2020optimization,Panda2024DTSemNet} have limited performance.

Due to the discrete nature of programs, search-based methods such as local search~\cite{marino2021programmatic,moraes2023choosing,aleixo2023show} and Monte Carlo Tree Search (MCTS)~\cite{gu2024pi} are well suited for generating candidate programs by applying production rules that extend or mutate existing programs. 
In domains where control flow is governed by fixed, discrete predicates (such as ``FrontIsClear'' in  Karel~\cite{trivedi2021learning}), search over program architecture alone can optimize effectively. In contrast, in continuous-control settings~\cite{verma2018programmatically}, control flow is often determined by parameterized conditions that map continuous features (such as velocities or angles) to boolean decisions.

However, these approaches are limited to gradient-free optimization due to the discreteness of programs, and they often fail to fully leverage collected experience, leading to poor sample efficiency. To address these limitations, $\pi$-PRL~\cite{qiu2022programmatic} relaxes discrete programs into continuous, differentiable derivation trees, by assigning probabilities to different program branches. This enables policy gradient methods to jointly optimize program architecture and predicate parameters. $\pi$-PRL has been extended by replacing its linear conditions with recurrent neural networks, improving generalization across tasks~\cite{Wu2024Synthesizing}. During training, this approach samples different tasks in each epoch and updates program-architecture weights and parameters sequentially.

A key challenge of using differentiable program relaxations is how to recover a final discrete policy. Post-hoc discretization can discard useful branches and parameters and lead to significant performance degradation, even if followed by additional fine-tuning. We present an example of such performance collapse in Sec.~\ref{sec:motivation}, highlighting the motivation for gradual discretization during training as opposed to post-hoc discretization after training.

\begin{wrapfigure}{r}{0.48\textwidth}
\vspace{-1em}
\centering
\begin{align*}
\text{Program}~E &:= A 
    \mid \text{ if } B \text{ then } A \text{ else } E \\
\text{Condition}~B &:= \phi_{\bm{w}}(s) > 0
\end{align*}
\caption{Domain-specific language (DSL) for programmatic policies. $A$ denotes a terminal action. $\phi_{\bm{w}}(s)$ is a parameterized predicate, specifically a linear function of state features.}
\label{fig:dsl}
\vspace{-1em}
\end{wrapfigure}

\section{Background}

\paragraph{Domain-specific language.} Following prior work~\cite{qiu2022programmatic}, we represent programmatic policies using the context-free domain-specific language (DSL) shown in Figure~\ref{fig:dsl}. The DSL specifies the syntactic space of programmatic policies by defining how programs are composed from three core components: \textit{perceptions} (state-dependent predicates), \textit{control flow} (branching), and \textit{actions} (terminal decisions).
A policy program $E$ is either a primitive action $A$ or a conditional statement of the form
\textit{if} $B$ \textit{then} $A$ \textit{else} $E$, where the condition $B$ is defined by a state predicate $\phi_{\bm{w}}(s) > 0$, parameterized by $\bm{w}$. We use linear functions of state features for $\phi_{\bm{w}}(s)$. Conditions determine choices between program branches, such as following the \texttt{if} or \texttt{else} branch. Given the DSL, a programmatic policy is constructed by starting from an initial (incomplete) program $E$ and iteratively expanding non-terminal nodes. This expansion process continues until all non-terminal nodes have been expanded, at which point the resulting program architecture is specified but still leaves predicates $\bm{w}$ to be chosen. We define a maximal program depth, at which we enforce the expansion of actions (terminal nodes) instead of additional non-terminals.

\paragraph{Markov decision processes.}
A Markov decision process (MDP)~\cite{sutton2018reinforcement}
is defined as a tuple $(\sS,\sA,\sR,\sP,\gamma)$, where $\sS$ is the set of states (e.g., velocity and angle in CartPole-v1), $\sA$ is the set of actions (e.g., left and right actions), $\sR:\sS\times \sA \times \sS \mapsto \mathbb{R}$ is the reward function, $\sP:\sS\times \sA \times \sS \mapsto [0,1]$ is the transition probability function, and $\gamma$ is the discount factor. 
A policy $\pi$ is a mapping from states to probability distributions over actions, where $\pi(a_t|s_t)$ is the probability of taking action $a_t$ in state $s_t$. 
The goal is to optimize a parameterized policy $\pi_\theta$ that maximizes the discounted cumulative reward, formulated as:
$ \max_{\theta}~J(\theta)=\mathbb{E}_{\tau \sim\pi_\theta}[\sum_{t=0}^{\infty} \gamma^t \sR(s_t,a_t,s_{t+1})],$ where $s_0$ is an initial state, and $\tau \sim\pi_{\theta}$ denotes a trajectory $(s_0,a_0,s_1,a_1,\dots, s_t,a_t,s_{t+1})$.

\section{Motivation: Failure mode of the post-hoc discretization }
\label{sec:motivation}

\begin{wrapfigure}{r}{0.55\textwidth}
    \vspace{-0.5em}
    \centering

    \includegraphics[width=0.9\linewidth]{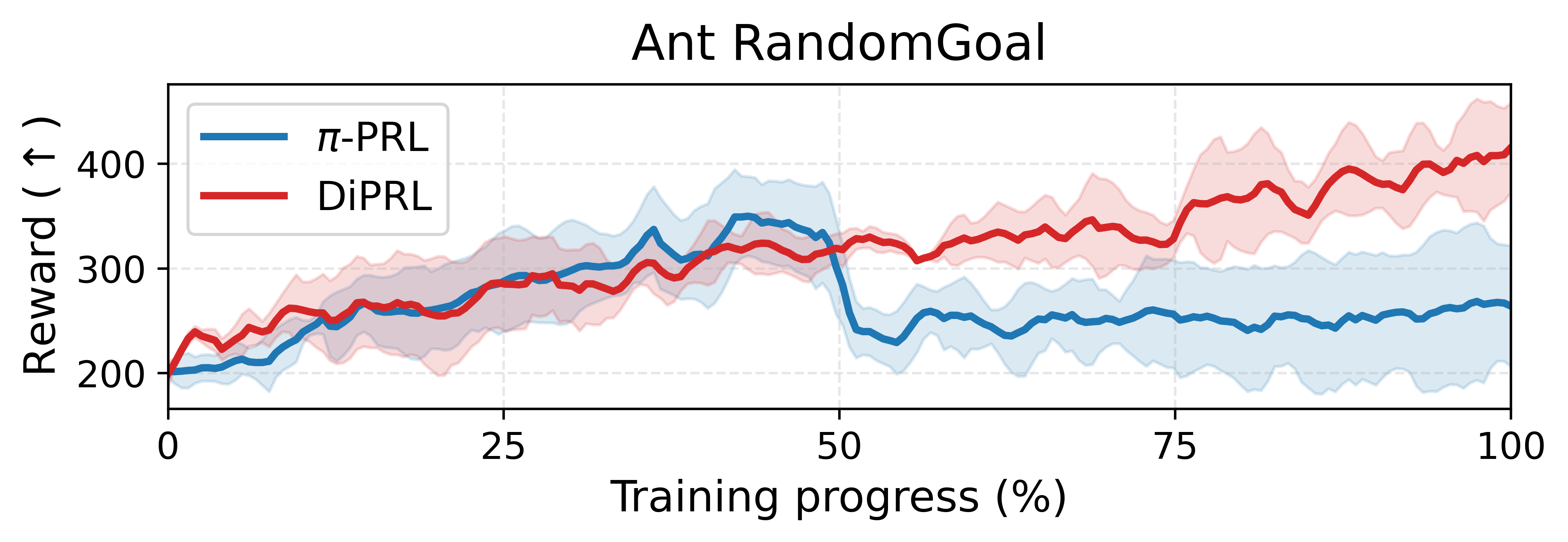}

    \caption{Motivating example of the performance drop caused by post-hoc discretization. }
    \label{fig:fail_example}
    \vspace{-0.5em}
\end{wrapfigure}

We present a failure mode of the post-hoc discretization in $\pi$-PRL, as shown in Fig.~\ref{fig:fail_example}. Both $\pi$-PRL and DiPRL perform similarly before 50\% training progress. However, after applying post-hoc discretization (following the original setting of $\pi$-PRL), its performance suddenly drops. Even after another 50\% of training, its performance is never recovered. The reason for this failure comes from the sudden step from the derivation tree that is trained via policy gradient, a parameterized distribution over multiple program architectures, to the eventually required output of a single discrete program. The post-hoc discretization used by $\pi$-PRL extracts this program using maximum likelihood among all possible programs, thus losing all learned contributions from other programs in the derivation tree. The extracted program loses too many program branches, which affects the expressivity of the policy. Even further fine-tuning therefore cannot recover the optimal performance.

\section{Differentiable Discrete Programmatic RL }
\begin{figure*}[t]
    \centering
    \includegraphics[width=\linewidth]{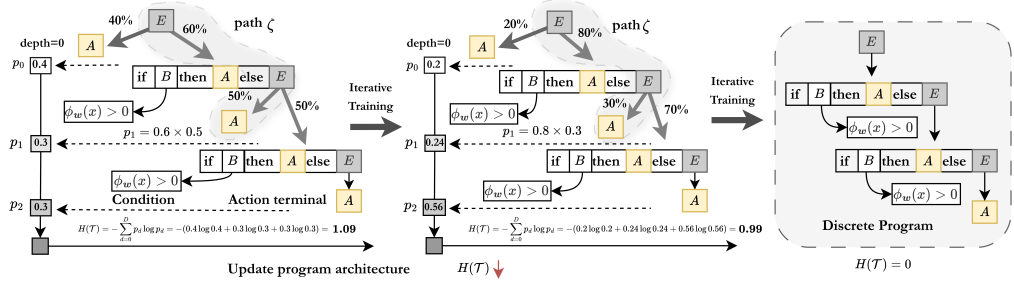}
    \caption{Workflow of DiPRL.
    $E$ denotes a program node. $A$ denotes an action node, marked in yellow. $B$ denotes a condition in an incomplete program \texttt{if} $B$ \texttt{then} $A$ \texttt{else} $E$.  The maximal depth $D_m$ is set to 2 in the figure. 
    A programmatic policy is relaxed into a continuous derivation program tree by fully expanding and assigning expansion probabilities to program branches. During the training, the program architecture entropy $H(\sT)$ decreases and encourages the relaxed architecture to approach a discrete program. }
    \label{fig:diPRL_example}
\end{figure*}

We motivate DiPRL by showing that applying post-hoc discretization to relaxed programmatic policies can lead to an unrecoverable performance drop even with fine-tuning (as presented in Fig.~\ref{fig:fail_example}). DiPRL mitigates this issue by introducing \emph{program architecture regularization} to $\pi$-PRL~\cite{qiu2022programmatic}, a continuous differentiable derivation tree. As shown in Figure~\ref{fig:diPRL_example}, the tree is built by recursively expanding non-terminal nodes according to the DSL in Figure~\ref{fig:dsl}. It represents a distribution over possible programs, so the final policy combines contributions from multiple program architectures. To encourage the relaxed tree to become discrete during training, DiPRL adds program architecture entropy as a regularizer and anneals it with automatic coefficient tuning, eliminating the need for a post-hoc training stage.

\subsection{Relaxation into differentiable derivation tree}
\label{sec:relaxation}

We follow the same procedure of $\pi$-PRL~\cite{qiu2022programmatic} to relax the program into a differential derivation tree. 
We denote the programmatic policy relaxed by the differentiable derivation tree as $\pi^\mathcal{T}_\mathcal{W}$, where $\mathcal{T}$ is the relaxed derivation tree used and $\mathcal{W}$ is the set of predicate and action parameters. The tree $\mathcal{T} = \{\mathcal{V}, \mathcal{E}\}$ consists of a node set $\mathcal{V}$ and an edge set $\mathcal{E}$. A node $v \in \mathcal{V}$ can be either a non-terminal, e.g., an \texttt{if-else} node, or a terminal action node. An edge $\langle v_i,v_j \rangle \in \mathcal{E}$ connects two nodes, and the expansion probability $P_{\mathcal{T}}(v_j\mid v_i)$ specifies how a non-terminal is expanded by the DSL. These expansion probabilities are state independent and induce a distribution over program paths. 

Let $\mathcal{P}$ be the set of program paths in the derivation tree. A program path $\zeta=\langle v_0,v_1,\cdots,v_n\rangle$ starts from the root and follows expansion decisions to a terminal program form. For each consecutive pair in the path, $v_{k+1}$ is a successor of $v_k$. The state-independent probability of $\zeta$ is the product of all expansions along the path:
$
    P_{\mathcal{T}}(\zeta)
    =
    \prod_{k=0}^{n-1}
    P_{\mathcal{T}}(v_{k+1}\mid v_k).
$

Predicate choices along a program path are state dependent. At an \texttt{if-else} node with predicate $\phi_{\bm{w}}(s_t)$, we relax the hard branch indicator $\mathds{1}[\phi_{\bm{w}}(s_t)>0]$ to $\sigma(\phi_{\bm{w}}(s_t))$, where $\sigma(x)=1/(1+e^{-x})$.
For a predicate-branch edge $v_i\rightarrow v_j$, let $b_{ij}=1$ for the \texttt{if} branch and $b_{ij}=0$ for the \texttt{else} branch. Its relaxed predicate-gating probability is $p_{ij}(s_t)=\sigma(\phi_{\bm{w}}(s_t))^{b_{ij}}(1-\sigma(\phi_{\bm{w}}(s_t)))^{1-b_{ij}}$.
Note that this probability is internal to an \texttt{if-else} node and determines whether \texttt{if} or \texttt{else} is taken, depending on the state. It is distinct from the program expansion probability $\text{Pr}(v_j^i\mid v_i)$, which determines how a non-terminal is expanded by the DSL (e.g., into an action node or another non-terminal node).

To enable policy gradient, we parameterize each action node with a parameterized action policy $\pi_{\bm{w}_\zeta}(a_t \mid s_t)$, specifically a categorical distribution over discrete actions or a Gaussian distribution for continuous actions. The state-dependent action distribution contributed by program path $\zeta$ is
\begin{equation}
    \pi_{\zeta,\mathcal{W}}(a_t\mid s_t)
    =
    \left(
    \prod_{\langle v_i,v_j\rangle\in\zeta}
    p_{ij}(s_t)
    \right)
    \pi_{\bm{w}_\zeta}(a_t\mid s_t).
\end{equation}
Then, the relaxed derivation tree that considers both state-independent architecture and state-dependent action distribution is denoted as follows:
\begin{equation}
    \pi^\mathcal{T}_\mathcal{W}(a_t \mid s_t)
    =
    \sum_{\zeta\in\mathcal{P}}
    P_{\mathcal{T}}(\zeta)
    \pi_{\zeta,\mathcal{W}}(a_t\mid s_t).
\end{equation}

\subsection{Program architecture regularization}
\label{sec:regularization}

While the relaxation described in Sec.~\ref{sec:relaxation} enables differentiability, it does not specify how to return a discrete policy after training. As discussed in Sec.~\ref{sec:motivation}, a significant performance drop may occur if the learned program remains a derivation tree and is discretized post-hoc, e.g., if a discrete program is extracted by selecting the most likely outgoing edge at each node based on $\max_{v_j} \Pr(v_j \mid v_i)$. Post-hoc discretization discards trained program branches whose contributions to performance may not be recoverable even with additional fine-tuning. To avoid this, we propose a novel regularization during training, gradually reducing the  \emph{uncertainty} of the program architecture.

\paragraph{Program architecture entropy}
Recall that the program expansion probabilities $\text{P}_{\mathcal{T}}(v_j\mid v_i)$ determine how each node in the derivation tree is expanded. These probabilities therefore characterize the uncertainty of the program architecture. A program becomes discrete when the expansion probabilities at each node collapse to a one-hot vector, so that only one expansion can be selected.

To quantify this uncertainty, we consider the architecture-dependent probability distribution over program paths, denoted by $\Pr(\zeta)$, where $\zeta$ is a complete path in the derivation tree. Let $\mathcal{P}$ denote the set of all possible paths. The generic program architecture entropy is defined as
\begin{equation}
    \hat{H}(\mathcal{T})
    =
    -\sum_{\zeta\in \mathcal{P}}
    \Pr(\zeta)\log \Pr(\zeta).
\end{equation}
Equivalently, if $P$ is the random variable representing the sampled program path, then
$\hat{H}(\mathcal{T}) = H(P)$. We then express this entropy in terms of program depth. Let $D$ be the random variable denoting the depth of a sampled path. Since the depth is determined by the path, we have $D=f(P)$. For a maximum depth $D_m$, define the probability of terminating at depth $d$ as
$p_d \triangleq \Pr(D=d)=\sum_{\zeta\in\mathcal{P}_d} \Pr(\zeta),\sum_{d=1}^{D_m} p_d = 1,$ where $\mathcal{P}_d$ is the set of all paths with depth $d$. The corresponding depth-based architecture entropy is
$H(\mathcal{T})=H(D)=-\sum_{d=1}^{D_m} p_d \log p_d.$ In the DSL used in this work, the depth uniquely determines the corresponding program path. Therefore, the conditional entropy term vanishes and the depth-based entropy is not merely a surrogate but exactly matches the generic path entropy, $H(\mathcal{T}) = \hat{H}(\mathcal{T})$.  We use this entropy as a regularizer during training to gradually reduce the uncertainty of the program architecture and encourage convergence toward a discrete program.

\paragraph{Regularized policy gradient}
High entropy indicates a larger uncertainty in the program architecture, while low entropy indicates high confidence in a particular architecture. By introducing this architecture regularization, we reformulate the MDP that maximizes the returns while converging to a discrete program as follows:
\begin{equation}
\max_{\mathcal{T}, \mathcal{W}}\;\; J(\mathcal{T}, \mathcal{W})\quad \text{s.t.} \quad H(\mathcal{T}) \le \bar{H},
\end{equation}
where $\bar{H}$ is the target architecture entropy. A fully discrete architecture would ideally have zero entropy, but setting $\bar{H}=0$ can force premature convergence to a poor program before useful branches have been explored. We therefore use a small positive target entropy and discuss its choice in Appendix~\ref{app:ablation_target}. Using a Lagrangian relaxation, we relax the objective:
\begin{equation}
\label{eq:joint_loss}
L{(\mathcal{T}, \mathcal{W},\alpha)} =\mathbb{E}_{\tau} \left[\sum_{t=0}^{T}\log \pi^\mathcal{T}_\mathcal{W}(a_t \mid s_t)\,\hat{A}_t\right]- \alpha (H(\mathcal{T})-\bar{H}),
\end{equation}
where $\alpha\ge0$, $ \tau \sim \pi_{\{\mathcal{T}, \mathcal{W}\}}$ and $\hat{A}_t$ denotes the advantage estimate that measures how good $a_t$ is in state $s_t$, following the policy gradient~\cite{schulman2017proximal}. We discuss strength of $\alpha$ in Appendix~\ref{app:regulari_ablation}.

Our regularization is different from entropy-based reinforcement learning methods such as Soft Actor-Critic (SAC)~\cite{haarnoja2018soft,wang2024negatively}. SAC encourages exploration by maximizing the entropy of the action distribution. However, our regularization operates directly on the program architecture, which aims at shaping the distribution over the derivation tree rather than over actions.

\subsection{Why post-hoc discretization can fail}
\label{sec:theory}

We provide a theoretical analysis (detailed in Appendix~\ref{app:theory_app}) for the failure mode of post-hoc program extraction. Let $\tilde{\mathcal{P}}\subseteq\mathcal{P}$ be the program paths retained after extracting the discrete program $\tilde{\mathcal{T}}$, and let $\tilde{\pi}$ denote the policy induced by this extracted program. For a program path $\zeta$, define the path-induced value:

\begin{equation}
    Q^\pi_\zeta(s)
    =
    \sum_{a\in\sA}\pi_{\zeta,\mathcal{W}}(a\mid s)Q^\pi(s,a),
\end{equation}

and let$m_{\mathcal{T}}=\sum_{\zeta\notin\tilde{\mathcal{P}}}P_{\mathcal{T}}(\zeta)$ be the deleted program-path mass. The performance difference between $\pi$ and $\tilde{\pi}$ is

\begin{equation}
    J(\tilde{\pi})-J(\pi)
    =
    \frac{1}{1-\gamma}\mathbb E_{s\sim \tau^{\tilde{\pi}}}\left[
    \sum_{\zeta\in\tilde{\mathcal{P}}}\left(P_{\tilde{\mathcal{T}}}(\zeta)-P_{\mathcal{T}}(\zeta)\right)Q^\pi_\zeta(s)
    -
    \sum_{\zeta\notin\tilde{\mathcal{P}}}P_{\mathcal{T}}(\zeta)Q^\pi_\zeta(s)
    \right].
\end{equation}

Let $Q_{\mathrm{del}}(s)$ be the average $Q^\pi_\zeta(s)$ value of deleted program paths, and let $Q_{\mathrm{keep}}(s)$ be the average value of retained program paths receiving the removed mass. Then

\begin{equation}
    J(\tilde{\pi})-J(\pi) = \frac{1}{1-\gamma}\mathbb E_{s\sim \tau^{\tilde{\pi}}}\left[m_{\mathcal{T}}\left(Q_{\mathrm{keep}}(s)-Q_{\mathrm{del}}(s)\right)\right].
\end{equation}
Suppose there exists a $\kappa(s)>0$, such that $Q_{\mathrm{del}}(s)-Q_{\mathrm{keep}}(s)\ge \kappa(s)$, then

\begin{equation}
    J(\tilde{\pi})-J(\pi)
    \le
    -\frac{1}{1-\gamma}
    \mathbb E_{s\sim \tau^{\tilde{\pi}}}[m_{\mathcal{T}}\kappa(s)].
\end{equation}

Thus, the discretization drop is affected by the amount of deleted program-path mass $m_{\mathcal{T}}$ and the value gap between deleted and retained program paths. DiPRL reduces this failure mode by making the program-path distribution concentrate during training.

\subsection{Why architecture entropy regularization helps}
The above section shows that post-hoc extraction is harmful when it removes probability mass from useful program paths. Architecture entropy regularization addresses this term directly. Let $p_{\max}=\max_{\zeta\in\mathcal{P}}P_{\mathcal{T}}(\zeta)$ be the probability of the program path selected during discretization, and let $m_{\mathcal{T}}=1-p_{\max}$ be the total path probability mass assigned to deleted paths. A low-entropy program-path distribution must concentrate on at least one path. Since Shannon entropy upper-bounds minimal entropy, $H(\mathcal{T}) \ge -\log p_{\max}$, and therefore $p_{\max}\ge\exp(-H(\mathcal{T}))$. Thus
\begin{equation}
    m_{\mathcal{T}}
    =
    1-p_{\max}
    \le
    1-\exp(-H(\mathcal{T}))
    \le
    H(\mathcal{T}).
\end{equation}
If the entropy regularizer drives $H(\mathcal{T})\le \epsilon, \epsilon>0$ before discretization, then the mass that can be discarded by selecting the maximum-probability path is at most $\epsilon$. Let $\Delta$ be a uniform upper bound on the retained-versus-deleted path-value gap, i.e.,
$|Q_{\mathrm{keep}}(s)-Q_{\mathrm{del}}(s)|\le\Delta$, then
\begin{equation}
\left|J(\tilde{\pi})-J(\pi)\right|
\le
\frac{\Delta}{1-\gamma}m_{\mathcal{T}}
\le
\frac{\Delta}{1-\gamma}\epsilon.
\end{equation}

Architecture entropy regularization does not assume that deleted paths have low value, instead, it reduces the total amount of program-path mass that can be deleted. When $H(\mathcal{T})$ is small, the relaxed derivation tree is already close to a single discrete program path in architecture space, so the policy evaluated after extraction is closer to the relaxed policy optimized during training.

\section{Experiments}
\label{sec:exp}
\paragraph{Domains}
We evaluate DiPRL and baselines on domains with discrete tasks, namely classic control tasks~\cite{brockman2016openai} including CartPole-v1, Acrobot-v1, and MountainCar-v0, as well as DoorKey and LunarLander-v2. We also compare algorithms on continuous tasks~\cite{qiu2022programmatic} like HalfCheetah Hurdle, Pusher2D, Ant RandomGoal and Ant CrossMaze.

\paragraph{Baselines}
We compare DiPRL with PPO~\cite{schulman2017proximal} as a neural policy baseline, VIPER~\cite{bastani2018verifiable}, and DTSemNets~\cite{Panda2024DTSemNet} as decision tree policy baselines, and $\pi$-PRL~\cite{qiu2022programmatic} as a differentiable programmatic policy baseline. In addition, we add $\pi_{\mathrm{cont.}}$-PRL, which denotes the program derivation tree trained by $\pi$-PRL before the post-hoc training stage. $\pi_{\mathrm{disc.}}$-PRL denotes the extracted discrete program from $\pi_{\mathrm{cont.}}$-PRL. $\pi$-PRL is the final programmatic policy after fine-tuning.

VIPER extracts a decision tree policy based on the neural policy trained by PPO, while DTSemNets directly trains a differentiable decision tree policy without an oracle. $\pi$-PRL~\cite{qiu2022programmatic} does not need an oracle, but requires post-hoc discretization and fine-tuning. The post-hoc discretization keeps the dominant branches according to the learned architecture distribution and replaces the relaxed sigmoid with a hard threshold. DiPRL does not need fine-tuning, and the final discretization is used only for evaluating the discrete program.
The maximal depth of DT~\cite{verma2018programmatically} and programs is set to 6, as suggested in \cite{qiu2022programmatic}. All algorithms are trained with the same budget on each task across three seeds. Implementation and experiment configuration are detailed in appendix~\ref{app:implement}.

\subsection{Discrete tasks}

\begin{table*}[t]
\centering
\caption{Rewards (mean $\pm$ std) on discrete benchmark tasks across three runs; higher is better. $\pi_{\mathrm{cont.}}$-PRL, $\pi_{\mathrm{disc.}}$-PRL, and $\pi$-PRL denote the relaxed policy before discretization, the discretized policy before fine-tuning, and the final fine-tuned policy, respectively. The best performance among policies (DTSemNets, $\pi$-PRL and DiPRL) without oracle is bold for each tasks. }
\setlength{\tabcolsep}{1pt}
\begin{tabular}{lccccc}
\toprule
Algorithm & CartPole-v1 & Acrobot-v1 & MountainCar-v0 & DoorKey & LunarLander-V2 \\
\midrule
PPO & $500.00 \pm 0.00$ & $-62.57 \pm 0.56$ & $-98.57 \pm 0.23$ & $0.97 \pm 0.00$ & $283.11 \pm 1.14$ \\
VIPER (PPO) & $500.00 \pm 0.00$ & $-67.83 \pm 3.63$ & $-97.13 \pm 2.08$ & $0.55 \pm 0.17$ & $164.33 \pm 54.25$ \\
\midrule
DTSemNets & $326.76 \pm 119.77$ & $-78.98 \pm 0.68$ & $-199.99 \pm 0.07$ & $0.00 \pm 0.00$ & $257.61 \pm 8.92$ \\
$\pi_{\mathrm{cont.}}$-PRL & $\mathbf{500.00 \pm 0.00}$ & $-80.31 \pm 2.12$ & $-172.50 \pm 38.89$ & $0.63 \pm 0.44$ & $235.72 \pm 19.50$ \\
$\pi_{\mathrm{disc.}}$-PRL & $\mathbf{500.00 \pm 0.00}$ & $-86.63 \pm 2.67$ & $-178.58 \pm 30.30$ & $0.32 \pm 0.32$ & $239.90 \pm 30.28$ \\
$\pi$-PRL & $\mathbf{500.00 \pm 0.00}$ & $-89.37 \pm 4.83$ & $-171.83 \pm 39.83$ & $0.66 \pm 0.39$ & $257.21 \pm 4.63$ \\
DiPRL (ours) & $\mathbf{500.00 \pm 0.00}$ & $-79.93 \pm 3.37$ & $\mathbf{-110.79 \pm 4.07}$ & $\mathbf{0.95 \pm 0.01}$ & $\mathbf{260.21 \pm 13.61}$ \\
\bottomrule
\end{tabular}

\label{tab:discrete_results}
\end{table*}

\begin{figure}[b]
    \centering
    \includegraphics[width=\linewidth]{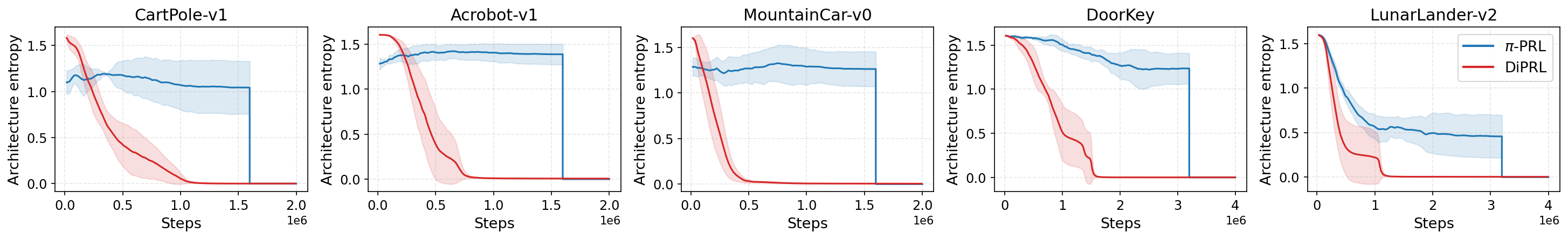}
    \caption{Architecture entropy curves in discrete tasks. The architecture entropy of $\pi$-PRL is forced to zero only via post-hoc discretization. DiPRL discretizes the derivation tree relaxation over time.}
    \label{fig:discrete_entropy_grid}
\end{figure}
Table~\ref{tab:discrete_results} summarizes the averaged rewards of DiPRL and the baselines on discrete tasks. We include PPO as a neural-policy reference, since it achieves strong performance across all environments. VIPER extracts decision-tree policies from a trained neural policy oracle; in our experiments, we use the PPO agent as this oracle. Although VIPER matches PPO on CartPole-v1 and performs well on Acrobot-v1 and MountainCar-v0, its performance is less stable on more challenging tasks, reflecting the distillation gap commonly observed in imitation-based policy extraction~\cite{qiu2022programmatic}. In particular, VIPER achieves only 0.55 on DoorKey and 164.33 on LunarLander-v2, despite access to PPO policies that obtain 0.97 and 283.11, respectively.

DTSemNets uses a specialized policy architecture that is semantically equivalent to an oblique decision tree. However, its architecture must be manually specified during training. We use the default policy setting from~\citet{Panda2024DTSemNet}. As shown in Table~\ref{tab:discrete_results}, DTSemNets performs competitively on Acrobot-v1 and LunarLander-v2, but it underperforms on CartPole-v1, MountainCar-v0, and DoorKey. For example, it obtains only 326.76 on CartPole-v1, collapses to nearly the minimum return on MountainCar-v0 with -200.19 , and fails to solve DoorKey, achieving 0.00. 

\begin{wrapfigure}{r}{0.5\textwidth}
    \centering
    \includegraphics[width=\linewidth]{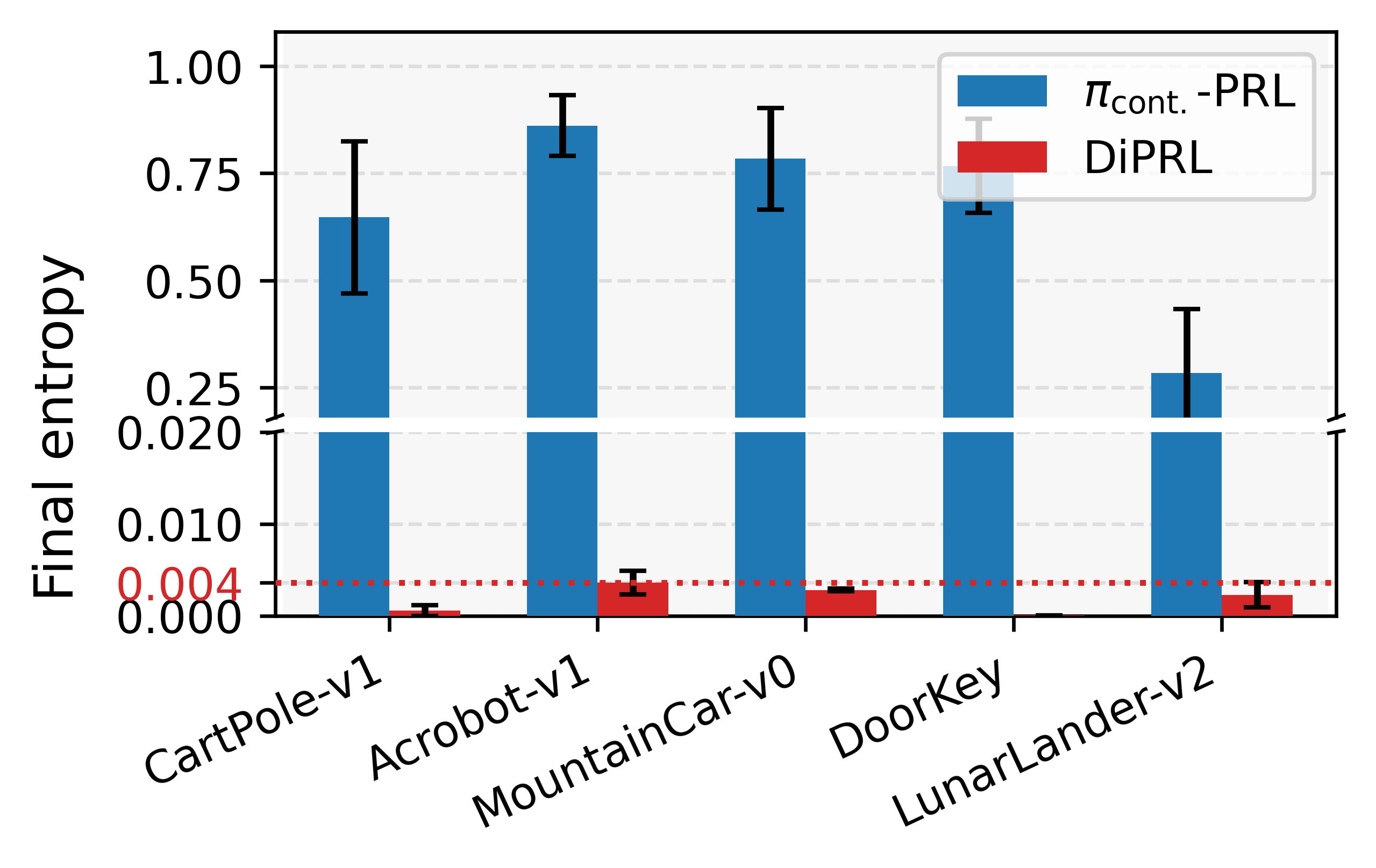}
    \caption{Normalized architecture entropy before discretization on discrete tasks. Lower values indicate architectures closer to discrete programs.}
    \label{fig:final_arch_entropy_discrete}
\end{wrapfigure}
The most closely related method to DiPRL is $\pi$-PRL, which also optimizes a continuous relaxation of a programmatic policy. However, the post-hoc discretization stage required by $\pi$-PRL can destabilize the final discrete policy by discarding useful learned branches. This effect is visible on DoorKey, where the continuous relaxed policy $\pi_{\mathrm{cont.}}$-PRL achieves 0.63, but the discretized $\pi_{\mathrm{disc.}}$-PRL policy drops to 0.32. DiPRL mitigates this issue by driving the relaxed architecture close to a discrete program during training, achieving 0.95 on DoorKey. We present the architecture entropy curve in Fig.~\ref{fig:discrete_entropy_grid}, where $\pi$-PRL still has a large entropy before the post-hoc discretization. Fig.~\ref{fig:final_arch_entropy_discrete} also shows that DiPRL maintains very small entropy, e.g., 0.004, whereas $\pi_{\mathrm{cont.}}$-PRL still has a large entropy $>0.25$ before the post-hoc discretization.

This result aligns with our theoretical analysis. $\pi$-PRL keeps a large $m_{\mathcal{T}}$, which aggravates the instability of the policy and may lead to a performance drop. Our DiPRL encourages gradual discretization over the course of the training, where DiPRL's architecture entropy eventually converges to zero, i.e., reducing $m_{\mathcal{T}}$. DiPRL allows alternative program branches to be gradually pruned during training, without requiring a post-hoc training stage. As a result, DiPRL avoids the significant performance drop typically observed in $\pi$-PRL, which even additional fine-tuning cannot always alleviate.

\subsection{Continuous tasks}

\begin{figure}[h]
    \centering
    \includegraphics[width=1\linewidth]{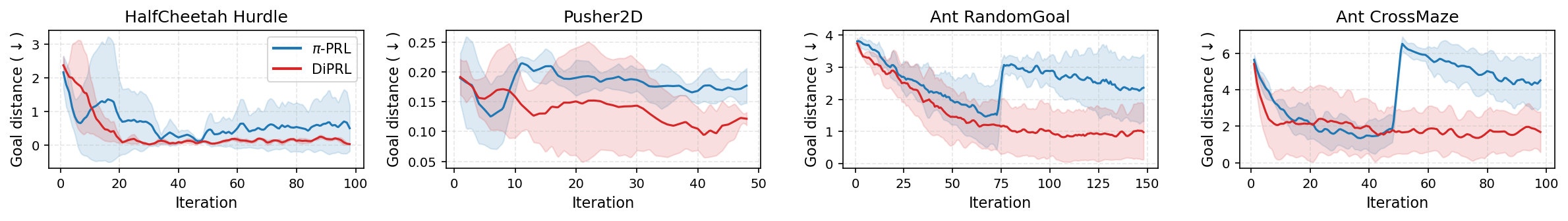}
    \caption{Goal distance curves in continuous tasks in a comparison with $\pi$-PRL and DiPRL; smaller is better. $\pi$-PRL collapses on all tasks except Pusher2D and its performance hardly recovers.}
    \label{fig:continuous_goal_distance}
\end{figure}

For the continuous tasks introduced by~\cite{qiu2022programmatic}, we report both episodic reward and goal distance; higher reward and lower goal distance indicate better performance. Table~\ref{tab:mujoco} summarizes the averaged rewards of DiPRL and the baselines, and Fig.~\ref{fig:continuous_goal_distance} presents the goal-distance curves. These continuous tasks are known to be challenging for vanilla PPO~\cite{qiu2022programmatic}. The relaxed policy $\pi_{\mathrm{cont.}}$-PRL performs well, especially on Ant CrossMaze, where it obtains 363.44 compared with 220.25 for DiPRL. However, $\pi$-PRL shows severe performance drop after discretization. On HalfCheetah Hurdle and Ant CrossMaze, its reward drops from 250.36 to -848.84 and from 363.44 to -506.32, respectively. Even with further fine-tuning, $\pi$-PRL hardly recovers from this drop, as also shown in Fig.~\ref{fig:continuous_goal_distance}. In contrast, DiPRL remains stable and achieves the best final discrete programmatic policy performance.

\begin{table*}[b]
    \centering
        \caption{Rewards (mean $\pm$ std) on the continuous tasks across three runs; higher is better.}
            \setlength{\tabcolsep}{3pt}
\begin{tabular}{lcccc}
\toprule
Algorithm & HalfCheetah Hurdle & Pusher2D & Ant RandomGoal & Ant CrossMaze \\
\midrule
PPO & $-780.43 \pm 758.89$ & $-87.07 \pm 2.77$ & $174.28 \pm 31.39$ & $-657.03 \pm 64.87$ \\
VIPER (PPO) & $-1107.49 \pm 928.25$ & $-87.85 \pm 2.79$ & $328.89 \pm 13.37$ & $-996.54 \pm 55.43$ \\
\midrule
DTSemNets & $-4790.10 \pm 22.09$ & $-79.13 \pm 4.80$ & $313.47 \pm 17.84$ & $-1010.04 \pm 37.50$ \\
$\pi_{\mathrm{cont.}}$-PRL & $250.36 \pm 478.97$ & $-82.44 \pm 2.84$ & $336.56 \pm 45.56$ & $\mathbf{363.44 \pm 180.23}$ \\
$\pi_{\mathrm{disc.}}$-PRL & $-848.84 \pm 626.73$ & $-102.65 \pm 0.66$ & $237.05 \pm 24.48$ & $-506.32 \pm 105.49$ \\
$\pi$-PRL & $-443.80 \pm 915.83$ & $-80.36 \pm 1.48$ & $264.75 \pm 61.80$ & $-223.58 \pm 226.08$ \\
DiPRL (ours) & $\mathbf{723.88 \pm 52.00}$ & $\mathbf{-78.62 \pm 1.59}$ & $\mathbf{413.12 \pm 47.62}$ & $220.25 \pm 75.94$ \\
\bottomrule
\end{tabular}
    \label{tab:mujoco}
\end{table*}

\subsection{Ablation study on regularization strength }

\begin{figure}[h]
    \centering
    \includegraphics[width=1\linewidth]{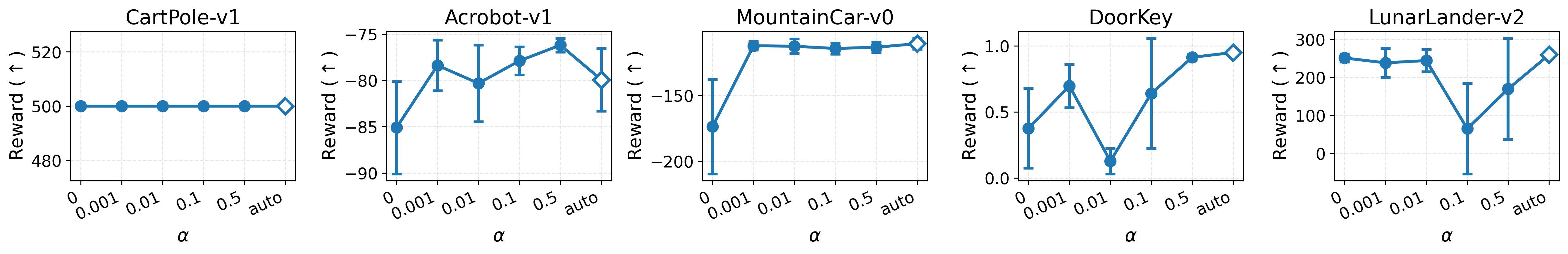}
    \caption{Ablation on regularizer $\alpha$ selected from set $\{0, 0.001, 0.01, 0.1, 0.5\}$ and automatic tuning. Automatic tuning (marked as rhombus $\textcolor{mplblue}{\diamond}$) is overall stable.}
    \label{fig:Ablation_regularization}
\end{figure}

 There is a trade-off in the choice of $\alpha$: a small coefficient may leave the final policy too relaxed, while a large coefficient may force premature convergence to a poor program. We evaluate the effect of $\alpha$ by comparing fixed coefficients $\alpha \in \{0, 0.001, 0.01, 0.1, 0.5\}$ with automatic tuning, which adapts $\alpha$ during training. All $\alpha$ settings reach optimal performance in CartPole-v1, as shown in Fig.~\ref{fig:Ablation_regularization}. On harder tasks, fixed coefficients show clear weaknesses. Small coefficients can leave useful behavior spread across multiple program branches, so final discretization removes part of the learned policy. Large coefficients can reduce this discretization gap, but they also push the architecture toward a discrete program too early and limit exploration. As a result, no fixed $\alpha$ is reliable across all tasks. Automatic tuning is more stable because it keeps architecture exploration flexible early in training and increases the pressure toward a discrete program later. Detailed results are reported in Appendix~\ref{app:regulari_ablation}.

\section{Conclusion}
\label{sec:con}
In this work, we first demonstrate a performance drop that can occur when extracting discrete programs post-hoc from a continuous program derivation tree in state-of-the-art PRL approaches. Even with further fine-tuning, it might be unrecoverable. We also theoretically analyze the failure mode.
To address the issue, we propose Differentiable Discrete Programmatic Reinforcement Learning (\textbf{DiPRL}) by introducing program architecture regularization. DiPRL does not require a separate post-hoc training stage. DiPRL uses gradient-based optimization over both program architecture and parameters. Instead of extracting a discrete program from a continuous relaxation after training, DiPRL integrates a program architecture entropy regularization that enables gradual discretization during training. The differentiable continuous derivation tree gradually converges to a discrete program, thereby mitigating the performance drop caused by the post-hoc discretization. 
Experiments on discrete and continuous tasks demonstrate the strong performance of DiPRL compared to differentiable decision tree policy and programmatic policy baselines. We also present an ablation study to verify the strength of the regularization.

\section{Limitation and future work}
\label{sec:limitation}
Programmatic policies are designed to be human-readable programs. It is important to verify if the derived program by $\pi$-PRL and DiPRL indeed improves interpretability. We leave the human-centered evaluation and study as important future work. It is interesting that DiPRL discovers strong policies grounded in a simple DSL with only \texttt{if-else} control flow. If an optimal strategy requires constructs this DSL lacks, such as loops and temporal operators for long-horizon tasks, DiPRL may converge to an interpretable but suboptimal program. In future work, we want to extend the DSL with temporal/iterative structure to better capture repetitive behaviors. We will also explore DiPRL on real-world problems such as job-shop scheduling.

\bibliography{main}
\bibliographystyle{unsrtnat}

\newpage
\appendix

\section{Broad Impact}
\label{app:bi}
Our DiPRL extends differential program derivation tree for training an interpretable programmatic policy with mitigating the performance drop of the post-hoc discretizations. A learned programmatic policy helps human users understand why an agent chooses an action in a given state. Besides, it can support policy debugging and failure detection before deployment. However, we note that a short program can still produce undesirable behavior, given biased states, or fail outside the training distribution. Over-trusting a policy is also a potential risk because its syntax is readable, even when its behavior has not been validated. These concerns are especially important if programmatic policies are used in real-world control, scheduling, robotics, or other decision-making systems.

\section{Theoretical analysis on the failure mode of the post-hoc discretization}
\label{app:theory_app}
We provide a theoretical analysis of the failure mode of post-hoc discretization while using program paths as the basic object. Let $\pi=\pi^\mathcal{T}_\mathcal{W}$ be the relaxed policy before discretization. Let $\mathcal{P}$ be the set of program paths and let $P_{\mathcal{T}}(\zeta)$ be the state-independent probability of program path $\zeta\in\mathcal{P}$. The policy contribution induced by program path $\zeta$ is
\begin{equation}
    \pi_{\zeta,\mathcal{W}}(a\mid s)
    =
    \left(
    \prod_{\langle v_i,v_j\rangle\in\zeta  }
    p_{ij}(s)
    \right)
    \pi_{\bm w_\zeta}(a\mid s),
\end{equation}
and the relaxed derivation-tree policy is
\begin{equation}
    \pi(a\mid s)
    =
    \sum_{\zeta\in\mathcal{P}}
    P_{\mathcal{T}}(\zeta)
    \pi_{\zeta,\mathcal{W}}(a\mid s).
\end{equation}

Let $\tilde{\mathcal{P}}\subseteq\mathcal{P}$ be the program paths retained after extracting the discrete program $\tilde{\mathcal{T}}$. The extracted policy $\tilde{\pi}$ uses the retained architecture distribution $P_{\tilde{\mathcal{T}}}$:
\begin{equation}
    \tilde{\pi}(a\mid s)
    =
    \sum_{\zeta\in\tilde{\mathcal{P}}}
    P_{\tilde{\mathcal{T}}}(\zeta)
    \pi_{\zeta,\mathcal{W}}(a\mid s),
    \qquad
    P_{\tilde{\mathcal{T}}}(\zeta)=0\;\text{for}\;\zeta\notin\tilde{\mathcal{P}}.
\end{equation}

For any policy $\mu$, define the discounted visited-state distribution as
\begin{equation}
    d^\mu(s)=(1-\gamma)\sum_{t=0}^{\infty}\gamma^t\Pr(s_t=s\mid \mu).
\end{equation}
By the performance-difference identity,
\begin{equation}
    J(\tilde{\pi})-J(\pi)
    =
    \frac{1}{1-\gamma}
    \mathbb E_{s\sim   \tau^{\tilde{\pi}}}
    \left[
        \sum_{a\in\sA}
        \left(\tilde{\pi}(a\mid s)-\pi(a\mid s)\right)
        Q^\pi(s,a)
    \right].
\end{equation}
Define the path-induced expected value under the reference value function $Q^\pi$ as
\begin{equation}
    Q^\pi_\zeta(s)
    =
    \sum_{a\in\sA}
    \pi_{\zeta,\mathcal{W}}(a\mid s)Q^\pi(s,a).
\end{equation}
Then,
\begin{equation}
    J(\tilde{\pi})-J(\pi)
    =
    \frac{1}{1-\gamma}
    \mathbb E_{s\sim   \tau^{\tilde{\pi}}}
    \left[
        \sum_{\zeta\in\tilde{\mathcal{P}}}
        \left(P_{\tilde{\mathcal{T}}}(\zeta)-P_{\mathcal{T}}(\zeta)\right)
        Q^\pi_\zeta(s)
        -
        \sum_{\zeta\notin\tilde{\mathcal{P}}}
        P_{\mathcal{T}}(\zeta)Q^\pi_\zeta(s)
    \right].
\end{equation}

We define the deleted architecture mass as
\begin{equation}
    m_{\mathcal{T}}
    =
    \sum_{\zeta\notin\tilde{\mathcal{P}}}
    P_{\mathcal{T}}(\zeta).
\end{equation}
Since $P_{\mathcal{T}}(\cdot)$ and $P_{\tilde{\mathcal{T}}}(\cdot)$ are both probability distributions over program paths, the total mass removed from deleted paths equals the total extra mass assigned to retained paths:
\begin{equation}
    m_{\mathcal{T}}
    =
    \sum_{\zeta\in\tilde{\mathcal{P}}}
    \left(P_{\tilde{\mathcal{T}}}(\zeta)-P_{\mathcal{T}}(\zeta)\right).
\end{equation}

Let $Q_{\mathrm{del}}(s)$ be the average $Q^\pi_\zeta(s)$ value of deleted program paths:
\begin{equation}
    Q_{\mathrm{del}}(s)
    =
    \frac{
        \sum_{\zeta\notin\tilde{\mathcal{P}}}
        P_{\mathcal{T}}(\zeta)Q^\pi_\zeta(s)
    }{m_{\mathcal{T}}}.
\end{equation}
Let $Q_{\mathrm{keep}}(s)$ be the average value of retained program paths receiving the extra probability mass:
\begin{equation}
    Q_{\mathrm{keep}}(s)
    =
    \frac{
        \sum_{\zeta\in\tilde{\mathcal{P}}}
        \left(P_{\tilde{\mathcal{T}}}(\zeta)-P_{\mathcal{T}}(\zeta)\right)
        Q^\pi_\zeta(s)
    }{m_{\mathcal{T}}}.
\end{equation}
Then the performance difference can be rewritten as
\begin{equation}
   J(\tilde{\pi})-J(\pi)
   =
   \frac{1}{1-\gamma}
   \mathbb E_{s\sim   \tau^{\tilde{\pi}}}
   \left[
        m_{\mathcal{T}}\left(Q_{\mathrm{keep}}(s)-Q_{\mathrm{del}}(s)\right)
   \right].
\end{equation}

This shows that post-hoc extraction decreases performance when the deleted program paths are more valuable than the retained program paths receiving their probability mass. More precisely, suppose there exists $\kappa(s)>0$ such that $Q_{\mathrm{del}}(s)-Q_{\mathrm{keep}}(s)\ge\kappa(s).$ Then
\begin{equation}
    J(\tilde{\pi})-J(\pi)
    \le
    -\frac{1}{1-\gamma}
    \mathbb E_{s\sim   \tau^{\tilde{\pi}}}
    \left[m_{\mathcal{T}}\kappa(s)\right].
\end{equation}

Finally, if the extracted program path is the maximum-probability path, $p_{\max}=\max_{\zeta\in\mathcal{P}}P_{\mathcal{T}}(\zeta)$ and $m_{\mathcal{T}}=1-p_{\max}$. Since Shannon entropy upper-bounds min-entropy,
\begin{equation}
    H(\mathcal{T})\ge -\log p_{\max},
    \qquad
    m_{\mathcal{T}}
    \le
    1-\exp(-H(\mathcal{T}))
    \le
    H(\mathcal{T}).
\end{equation}
Let $\epsilon$ be any upper bound on the achieved entropy before extraction, $H(\mathcal{T})\le\epsilon$. It can be the target $\bar{H}$. Let $\Delta$ be a uniform bound, $|Q_{\mathrm{keep}}(s)-Q_{\mathrm{del}}(s)|\le\Delta$ for states under $  \tau^{\tilde{\pi}}$. Then
\begin{equation}
    \left|J(\tilde{\pi})-J(\pi)\right|
    \le
    \frac{\Delta}{1-\gamma}m_{\mathcal{T}}
    \le
    \frac{\Delta}{1-\gamma}\epsilon.
\end{equation}
DiPRL targets the post-hoc extraction failure mode by making the architecture distribution concentrate during training, so that the deleted path mass $m_{\mathcal{T}}$ is small before the final discrete program is evaluated.

\section{Implementation details}
\label{app:implement}

This section summarizes the implementation and training settings used for the methods reported in the main tables. All aggregate results use three random seeds, $\{123,321,456\}$, and report the mean and standard deviation across seeds under 30 evaluation episodes.

\subsection{$\pi$-PRL}
We train $\pi$-PRL with the program derivation relaxation used by the original implementation~\cite{qiu2022programmatic}. During Phase 1, the policy alternates between architecture search and program-parameter optimization. The architecture step updates the distribution over candidate program depths, while the program step updates the predicates and action weights inside the candidate programs. 

\paragraph{Post-hoc discretization and fine-tuning in $\pi$-PRL.}
After Phase 1, $\pi$-PRL converts the  program derivation relaxation into a single discrete program. Concretely, the learned architecture distribution is frozen and the most likely program depth is selected by argmax over the architecture probabilities. The selected program then copies the corresponding learned predicates and action weights from the trained fusion program. We report this extracted policy before additional training as $\pi_{\mathrm{disc}}$-PRL. Standard $\pi$-PRL~\cite{qiu2022programmatic} then performs a Phase-2 fine-tuning stage on this fixed extracted program. The architecture is no longer searched, but the program parameters and value baseline continue to be optimized. Phase 2 uses $20\%$ of the total training budget for discrete tasks. For continuous tasks, we use $50\%$ of the total budget on HalfCheetah Hurdle, Ant RandomGoal, and Ant CrossMaze, and $80\%$ on Pusher2D, following the original settings of $\pi$-PRL~\cite{qiu2022programmatic}.

\subsection{DiPRL}
DiPRL uses the same program-derivation-graph parameterization as $\pi$-PRL, but introduces architecture entropy regularization. $H(\mathcal{T})$ measures architecture uncertainty, which is determined by the maximal depth $D_m$, i.e., $H(\mathcal{T})=H(D)= -\sum_{d=1}^{D_m} p_d \log p_d$. Thus, $\log{D_m}$ provides the natural scale of the uncertainty. The target entropy is $0.1\log{D_m}$. DiPRL is trained as a single phase and skips the post-extraction Phase-2 fine-tuning used by $\pi$-PRL. The final policy is the extracted discrete program from this single phase run. DiPRL uses automatic tuning via the dual update for $\alpha$ with a learning rate $10^{-4}$. We
initialize $\log \alpha=-10$, i.e., $\alpha_{\text{initial}}=\exp(-10)$, and clamp $\log \alpha$ to $[-10,10]$ during training.

\subsection{PPO}
The PPO baselines use the PPO implementation of Stable-Baselines3~\cite{stable-baselines3}.
\begin{itemize}
    \item \textbf{Discrete tasks.} PPO is trained with 2M steps on classic control, while DoorKey and LunarLander are trained for 4M steps. These runs use a learning rate $10^{-3}$ and a minibatch size of 64.
    \item \textbf{Continuous tasks.}  PPO is trained for 250k steps for Pusher2D, 5M for HalfCheetah Hurdle, 6M for Ant RandomGoal, and 5M for Ant CrossMaze. These runs use a learning rate $10^{-3}$ and a minibatch size of 64.
\end{itemize}
Note that both $\pi$-PRL and DiPRL keep the same hyperparameter setting with PPO but replace the neural policy with a programmatic policy.

\subsection{VIPER}
VIPER~\cite{bastani2018verifiable} is trained as a post-hoc distillation baseline from the PPO oracle for the same task and seed. VIPER uses Q-Dagger with a decision-tree classifier of maximum depth 6 and 100k sampled timesteps per run.

\subsection{DTSemNets}
DTSemNets is trained using the original implementation~\cite{Panda2024DTSemNet}. Budgets for all tasks are set to be the same as DiPRL.

\subsection{Computational Resource}
All algorithms are trained and tested on a 128-CPU server.
\begin{itemize}
    \item CPU: AMD Rome 7H12 (2x) 64 Cores/Socket 2.6GHz 280W
    \item CPU memory: 256 GiB DRAM (2 GiB per core)
    \item DIMMs: 16 x 16GiB 3200MHz, DDR4 
\end{itemize}

\begin{table}[thb]
\caption{List of licenses for assets used in this work}
\label{tab:license}
\resizebox{\textwidth}{!}{
\begin{tabular}{lc p{0.2\linewidth} r}
\toprule
Resource & Type & Link & License \\
\midrule
Tianshou~\cite{weng2022tianshou} & Code & \url{https://github.com/thu-ml/tianshou/tree/master?tab=readme-ov-file} &MIT License\\
OpenAI'Gym& Code & \url{https://github.com/openai/gym} & Available for academic use\\
\citet{qiu2022programmatic} & Code & \url{https://github.com/RU-Automated-Reasoning-Group/pi-PRL} & Available for academic use\\
Stable-Baselines3~\cite{stable-baselines3} & Code &\url{https://github.com/DLR-RM/stable-baselines3} & Available for academic use\\
$\pi$-PRL& Code &\url{https://github.com/RU-Automated-Reasoning-Group/pi-PRL} & Available for academic use\\
DTSemNets~\cite{Panda2024DTSemNet}& Code&\url{https://github.com/CPS-research-group/dtsemnet}  & Available for academic use\\

\bottomrule
\end{tabular}
}
\end{table}

\clearpage

\section{Ablation study on target entropy}
\label{app:ablation_target}
\begin{figure}[h]
    \centering
    \includegraphics[width=1\linewidth]{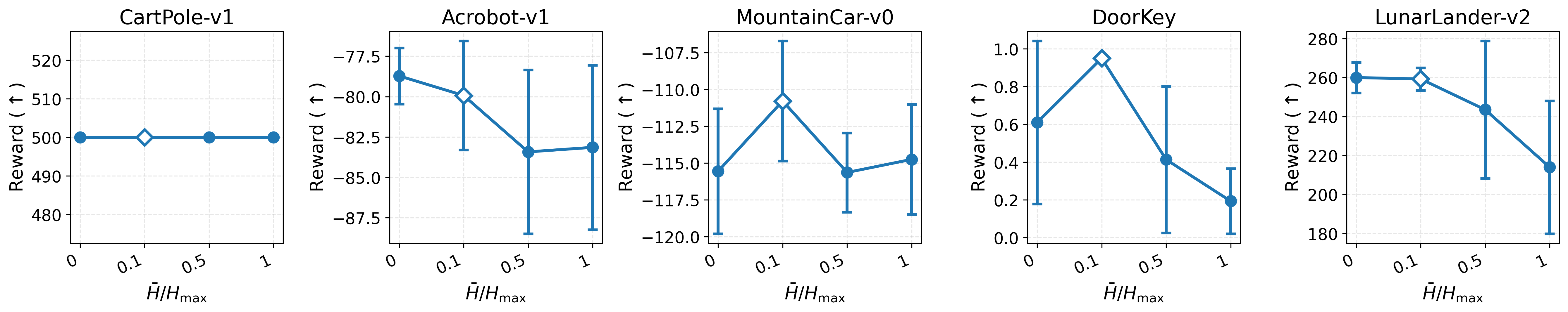}
    \caption{Ablation on target entropy selected from $\bar{H}\in\{0, 0.1\log{D_m}, 0.5\log{D_m}, \log{D_m}\}$.}
    \label{fig:ablation_target_app}
\end{figure}

There is a trade-off in setting $\bar{H}$. For example, setting $\bar{H}=0$ may cause premature collapse, resulting in a trivial program with limited expressivity. $H(\mathcal{T})$ measures architecture uncertainty, which is determined by the maximal depth $D_m$, i.e., $H(\mathcal{T})=H(D)= -\sum_{d=1}^{D_m} p_d \log p_d$. Thus, $\log{D_m}$ provides the natural scale of the uncertainty.
In preliminary experiments, we studied the effect of the target entropy by considering $\bar{H}\in\{0, 0.1\log{D_m}, 0.5\log{D_m}, \log{D_m}\}$. We consider $\log{D_m}$ as $H_{\mathrm{max}}$. Based on Tab.~\ref{tab:ablation_target_entropy} and Fig.~\ref{fig:ablation_target_app}, we found that the performance degenerates with larger target entropy. It is intuitive since large entropy has less effect in encouraging a discrete program, while zero might lead to a premature collapse. Thus, we select $0.1\log{D_m}$ as the target entropy for DiPRL.

\begin{table}[h]
    \centering
     \setlength{\tabcolsep}{1pt}
     \caption{Ablation on target entropy selected from $\bar{H}\in\{0, 0.1\log{D_m}, 0.5\log{D_m}, \log{D_m}\}$.}
\begin{tabular}{lccccc}
\toprule
Target entropy $\bar{H}$ & CartPole-v1 & Acrobot-v1 & MountainCar-v0 & DoorKey & LunarLander-v2 \\
\midrule
0 & $500.00 \pm 0.00$ & $-78.72 \pm 1.74$ & $-115.56 \pm 4.25$ & $0.61 \pm 0.43$ & $259.96 \pm 7.89$ \\
0.1$\log{D_m}$ & $500.00 \pm 0.00$ & $-79.93 \pm 3.37$ & $-110.79 \pm 4.07$ & $0.95 \pm 0.01$ & $260.21 \pm 13.61$ \\
0.5$\log{D_m}$ & $500.00 \pm 0.00$ & $-83.42 \pm 5.07$ & $-115.64 \pm 2.69$ & $0.41 \pm 0.39$ & $243.48 \pm 35.23$ \\
$\log{D_m}$ & $500.00 \pm 0.00$ & $-83.14 \pm 5.09$ & $-114.76 \pm 3.73$ & $0.19 \pm 0.17$ & $213.95 \pm 34.17$ \\
\bottomrule
\end{tabular}
    \label{tab:ablation_target_entropy}
\end{table}

\section{Ablation study on regularization strength}
\label{app:regulari_ablation}

 The coefficient $\alpha$ controls how strongly DiPRL penalizes architecture entropy. If $\alpha$ is too small, the relaxed policy may still rely on several program branches when the final program is extracted. If $\alpha$ is too large, the architecture may become discrete before the policy has explored useful branches and parameters.

\paragraph{Automatic coefficient tuning.}
It is hard to manually select a fixed penalty coefficient $\alpha$ for $H(\sT)$ (see Eq.~\ref{eq:joint_loss}) across different tasks. We therefore tune $\alpha$ during training. The dual objective with respect to $\alpha$ is defined as $g(\alpha)=\max_{\sT,\sW}\mathcal{L}(\sT,\sW,\alpha)$. We update $\alpha$ by approximating the dual gradient according to~\cite{boyd2004convex}:
\begin{equation}
\nabla_{\alpha} g(\alpha) = \bar{H} - H(\sT),
\end{equation} 

In practice, we optimize $\beta = \log \alpha$ to ensure positivity. This update increases $\alpha$ when the architecture entropy is above the target and decreases it when the entropy is already low enough. Thus, automatic tuning allows exploration in the early stage and encourages convergence to a discrete program later.

\paragraph{Ablation results.}

\begin{table}[t]
    \centering
    \small
    \setlength{\tabcolsep}{1pt}
    \caption{Ablation of the regularization coefficient $\alpha$ on discrete-control tasks. In the Discretization column, $\times$ denotes the relaxed policy before extraction and $\checkmark$ denotes the extracted discrete program. No fine-tuning is applied.}
    \label{tab:fixed_alpha_ablation}
    \begin{tabular}{llccccc}
        \toprule
        Algorithm & Discretization & CartPole-v1 & Acrobot-v1 & MountainCar-v0 & DoorKey & LunarLander-v2 \\
        \midrule
        \multirow{2}{*}{DiPRL$_{0}$}
            & $\times$     & $500.00 \pm 0.00$ & $-80.99 \pm 2.15$ & $-148.17 \pm 36.87$ & $0.63 \pm 0.45$ & $250.05 \pm 13.40$ \\
            & $\checkmark$ & $500.00 \pm 0.00$ & $-85.09 \pm 5.01$ & $-173.74 \pm 35.73$ & $0.38 \pm 0.30$ & $250.82 \pm 10.89$ \\
        \midrule
        \multirow{2}{*}{DiPRL$_{0.001}$}
            & $\times$     & $500.00 \pm 0.00$ & $-77.89 \pm 3.51$ & $-111.37 \pm 3.85$ & $0.94 \pm 0.00$ & $253.36 \pm 21.41$ \\
            & $\checkmark$ & $500.00 \pm 0.00$ & $-78.38 \pm 2.72$ & $-112.36 \pm 3.12$ & $0.70 \pm 0.16$ & $237.98 \pm 38.33$ \\
        \midrule
        \multirow{2}{*}{DiPRL$_{0.01}$}
            & $\times$     & $500.00 \pm 0.00$ & $-81.23 \pm 4.15$ & $-111.90 \pm 4.05$ & $0.62 \pm 0.44$ & $254.59 \pm 4.42$ \\
            & $\checkmark$ & $500.00 \pm 0.00$ & $-80.32 \pm 4.14$ & $-112.71 \pm 5.53$ & $0.13 \pm 0.10$ & $244.04 \pm 29.27$ \\
        \midrule
        \multirow{2}{*}{DiPRL$_{0.1}$}
            & $\times$     & $500.00 \pm 0.00$ & $-78.59 \pm 1.46$ & $-116.64 \pm 1.57$ & $0.92 \pm 0.02$ & $61.03 \pm 124.29$ \\
            & $\checkmark$ & $500.00 \pm 0.00$ & $-77.89 \pm 1.52$ & $-114.44 \pm 4.12$ & $0.64 \pm 0.42$ & $65.03 \pm 118.85$ \\
        \midrule
        \multirow{2}{*}{DiPRL$_{0.5}$}
            & $\times$     & $500.00 \pm 0.00$ & $-80.05 \pm 3.82$ & $-111.24 \pm 3.94$ & $0.92 \pm 0.03$ & $165.26 \pm 144.34$ \\
            & $\checkmark$ & $500.00 \pm 0.00$ & $-76.19 \pm 0.73$ & $-113.46 \pm 3.99$ & $0.91 \pm 0.02$ & $169.42 \pm 133.09$ \\
        \midrule
        \multirow{2}{*}{DiPRL$_{\mathrm{auto}}$}
            & $\times$     & \cellcolor{gray!15}$500.00 \pm 0.00$ & \cellcolor{gray!15}$-79.08 \pm 2.77$ &\cellcolor{gray!15} $-113.40 \pm 3.55$ & \cellcolor{gray!15}$0.95 \pm 0.01$ &\cellcolor{gray!15} $248.54 \pm 2.00$ \\
            & $\checkmark$ & \cellcolor{gray!15}$500.00 \pm 0.00$ & \cellcolor{gray!15}$-79.93 \pm 3.37$ & \cellcolor{gray!15}$-110.79 \pm 4.07$ & \cellcolor{gray!15}$0.95 \pm 0.01$ & \cellcolor{gray!15}$260.21 \pm 13.61$ \\
        \bottomrule
    \end{tabular}
\end{table}

Table~\ref{tab:fixed_alpha_ablation} compares each policy before and after discretization. The $\times$ rows evaluate the relaxed policy before extracting the final program, while the $\checkmark$ rows evaluate the extracted discrete program. No fine-tuning is applied. This comparison shows whether a choice of $\alpha$ already learns a stable discrete architecture during training or still depends on a relaxed mixture of program branches.

Small coefficients do not consistently control the discretization gap. For example, DiPRL$_{0.001}$ reaches $0.94$ on DoorKey before discretization, but drops to $0.70$ after extraction. DiPRL$_{0.01}$ shows an even larger DoorKey drop, from $0.62$ to $0.13$. DiPRL$_0$ also degrades on MountainCar-v0, from $-148.17$ to $-173.74$. These drops indicate that weak regularization can leave important behavior spread across several branches of the relaxed derivation tree. When the final program is extracted, some of these branches are removed, and the resulting discrete program is worse than the relaxed policy.

Large fixed coefficients reduce this discretization gap in some cases, but they can hurt learning before extraction. DiPRL$_{0.5}$ is stable on DoorKey after discretization, changing only from $0.92$ to $0.91$, but it performs poorly on LunarLander-v2 compared with the automatic setting. DiPRL$_{0.1}$ has the same problem: it remains far below the best LunarLander-v2 reward and still drops on DoorKey after discretization. This suggests that forcing discreteness too strongly can reduce the exploration of useful program branches and parameters.

DiPRL$_{\mathrm{auto}}$ avoids choosing a single fixed coefficient for all tasks. It keeps DoorKey unchanged after discretization at $0.95$, improves MountainCar-v0 after extraction from $-113.40$ to $-110.79$, and achieves the best LunarLander-v2 result after discretization. Automatic tuning keeps architecture exploration flexible early in training, then increases pressure toward a stable discrete program when the learned policy is ready to be extracted.

\clearpage
\section{Additional experiment results}

\subsection{Program depth comparison between $\pi$-PRL and DiPRL}
Tab~\ref{tab:main_prog_depth} presents the depth of the final discrete program of $\pi$-PRL and DiPRL. $\pi$-PRL has to discard branches, thus has few chance to explore deeper programs. DiPRL encourages discrete program convergence during training. Better programs with deeper depth can have higher chance to be found.

\begin{table}[h]
    \centering
        \caption{Final discrete program depth (avg. $\pm$ std) for $\pi$-PRL and DiPRL.}
\begin{tabular}{lcc}
\toprule
Task & $\pi$-PRL  & DiPRL (ours) \\
\midrule
CartPole-v1 & $4.00 \pm 0.82$ & $4.67 \pm 0.47$ \\
Acrobot-v1 & $2.33 \pm 0.47$ & $4.33 \pm 0.47$ \\
MountainCar-v0 & $1.33 \pm 0.47$ & $3.67 \pm 0.47$ \\
DoorKey-6x6 & $3.67 \pm 1.89$ & $4.67 \pm 0.47$ \\
LunarLander-v2 & $4.33 \pm 0.47$ & $4.33 \pm 0.47$ \\
\midrule
HalfCheetah Hurdle & $1.67 \pm 0.47$ & $1.67 \pm 0.94$ \\
Pusher2D & $1.67 \pm 0.47$ & $2.33 \pm 0.47$ \\
Ant RandomGoal & $3.00 \pm 1.41$ & $3.33 \pm 1.25$ \\
Ant CrossMaze & $2.33 \pm 1.89$ & $2.67 \pm 0.47$ \\
\bottomrule
\end{tabular}

    \label{tab:main_prog_depth}
\end{table}

\subsection{Architecture entropy comparison between $\pi_{\mathrm{cont.}}$-PRL and DiPRL}

Table~\ref{tab:final_arch_entropy} reports the final normalized architecture entropy before discretization. An interesting point is that $\pi$-PRL drops to a 0.004 architecture entropy in HalfCheetah Hurdle, but it still experiences a performance drop. This is because the reported architecture entropy measures the uncertainty of the program architecture. After extraction, $\pi$-PRL instantiates a new program policy for Phase 2 fine-tuning. The stochastic action distribution, considering the parameterized action terminal, is still unstable with hard thresholdings. Even when the architecture is nearly deterministic, the behavior of the discrete program may be identical to the relaxed policy. That is why $\pi$-PRL introduces the post-hoc fine-tuning. DiPRL consistently drives the architecture entropy close to zero while avoiding cases like this, whereas $\pi_{\mathrm{cont.}}$-PRL often remains high-entropy on almost discrete and continuous tasks. This supports the claim that DiPRL learns a near-discrete architecture during training, reducing the mismatch introduced by discretization.

\begin{table}[h]
\centering
\caption{Final normalized architecture entropy before discretization. Values are mean $\pm$ std across three seeds. Lower values indicate that the relaxed architecture is closer to a discrete program.}
\label{tab:final_arch_entropy}
\begin{tabular}{lcc}
\toprule
Task & $\pi_{\mathrm{cont.}}$-PRL & DiPRL \\
\midrule
CartPole-v1 & 0.647 $\pm$ 0.178 & 0.001 $\pm$ 0.001 \\
Acrobot-v1 & 0.861 $\pm$ 0.070 & 0.004 $\pm$ 0.001 \\
MountainCar-v0 & 0.784 $\pm$ 0.119 & 0.003 $\pm$ 0.000 \\
DoorKey & 0.767 $\pm$ 0.109 & 0.000 $\pm$ 0.000 \\
LunarLander-v2 & 0.284 $\pm$ 0.149 & 0.002 $\pm$ 0.001 \\
\midrule
HalfCheetah Hurdle & 0.004 $\pm$ 0.006 & 0.000 $\pm$ 0.000 \\
Pusher2D & 0.407 $\pm$ 0.328 & 0.000 $\pm$ 0.000 \\
Ant RandomGoal & 0.490 $\pm$ 0.071 & 0.000 $\pm$ 0.000 \\
Ant CrossMaze & 0.785 $\pm$ 0.063 & 0.024 $\pm$ 0.033 \\
\bottomrule
\end{tabular}
\end{table}

\subsubsection{Discrete tasks}
\begin{figure}
    \centering
    \includegraphics[width=0.9\linewidth]{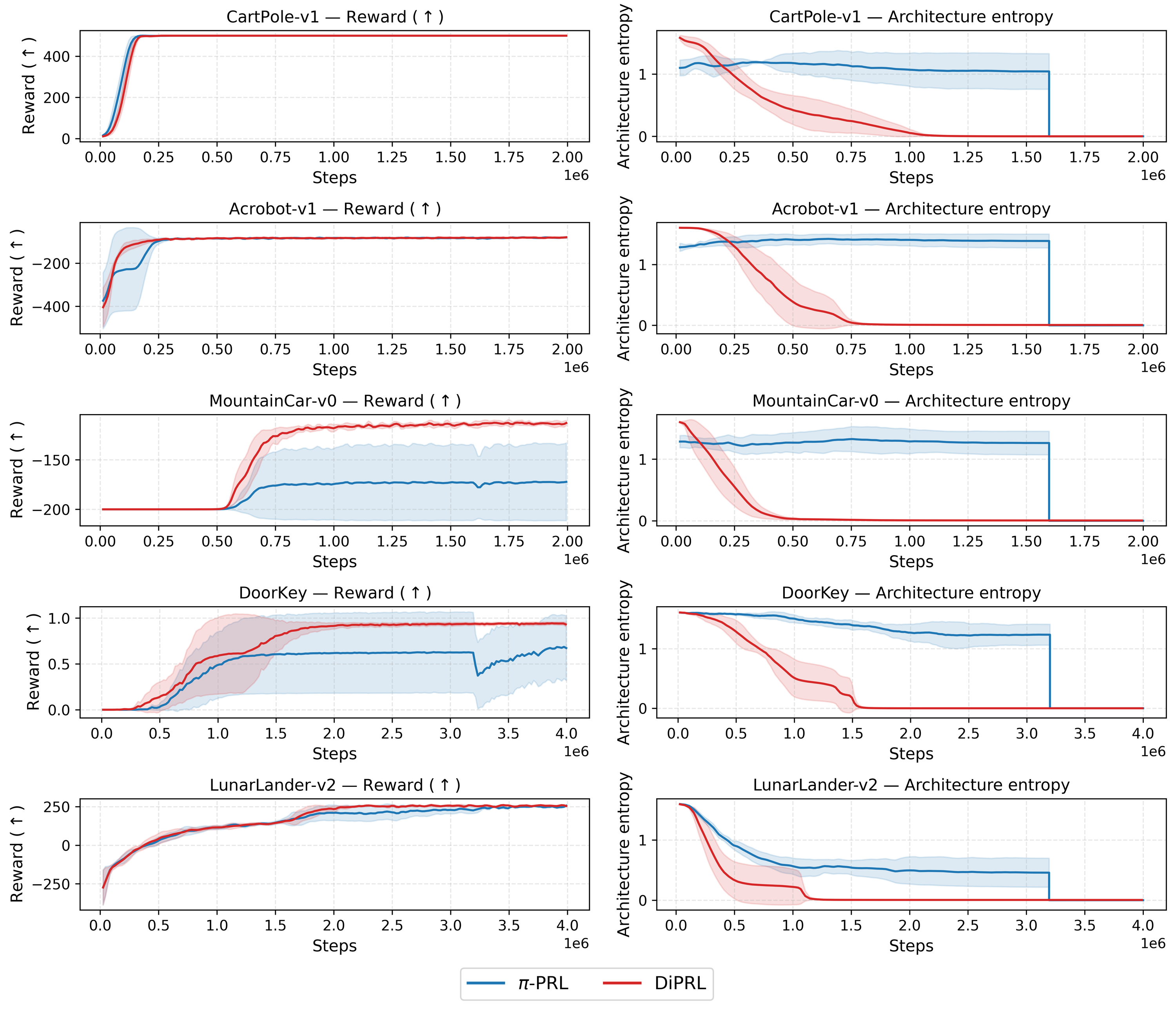}
    \caption{Training curves on discrete tasks. The left column reports reward, where higher is better, and the right column reports architecture entropy. DiPRL reduces architecture entropy during training while maintaining competitive reward.}
    \label{fig:discrete_grid}
\end{figure}

Fig.~\ref{fig:discrete_grid} shows that DiPRL keeps competitive reward while gradually driving the architecture entropy to zero. This trend is important because the final policy is a discrete program. Low entropy means that the learned derivation tree has already concentrated on one architecture before evaluation. In contrast, $\pi$-PRL often keeps high architecture entropy for most of training and only becomes discrete after the post-hoc extraction step. The reward curves are similar on simple tasks such as CartPole-v1 and Acrobot-v1, but DiPRL gives clearer gains on harder tasks such as MountainCar-v0 and DoorKey, where stable discretization is more important.

\subsubsection{Continuous tasks}
\begin{figure}
    \centering
    \includegraphics[width=0.9\linewidth]{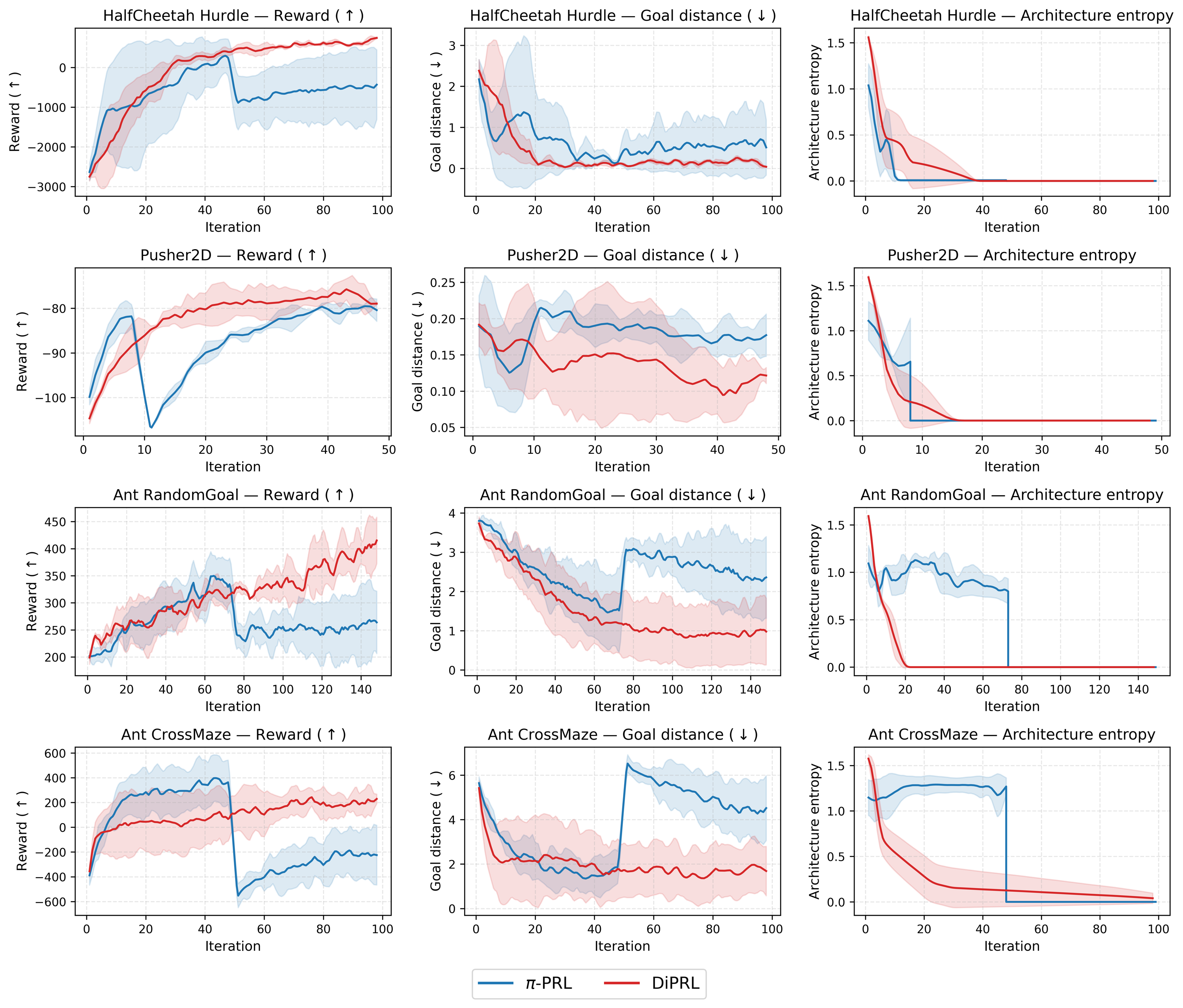}
    \caption{Training curves on continuous tasks. Each row reports reward, goal distance, and architecture entropy for one task. Higher reward and lower goal distance indicate better performance.}
    \label{fig:continuous_grid}
\end{figure}

Fig.~\ref{fig:continuous_grid} reports reward, goal distance, and architecture entropy on continuous-control tasks. Tab.~\ref{tab:app_continuous_distance} presents the goal distance. DiPRL generally improves reward while reducing goal distance, especially on HalfCheetah Hurdle and Ant RandomGoal. The entropy curves show that DiPRL also converges toward a discrete architecture during training. This supports the same conclusion as in the discrete tasks. The policy does not need a sudden post-hoc discretization step, because the architecture is already close to discrete when training ends.

\begin{table}[]
    \centering
        \caption{Goal distance curves in continuous tasks; smaller is better.}
\begin{tabular}{lcccc}
\toprule
Algorithm & HalfCheetah Hurdle & Pusher2D & Ant RandomGoal & Ant CrossMaze \\
\midrule
PPO & $0.53 \pm 0.72$ & $0.66 \pm 0.17$ & $3.80 \pm 0.48$ & $6.65 \pm 0.06$ \\
VIPER (PPO) & $0.60 \pm 0.56$ & $0.59 \pm 0.24$ & $2.93 \pm 0.33$ & $8.17 \pm 0.74$ \\
\midrule
DTSemNets & $10.27 \pm 0.69$ & $0.14 \pm 0.06$ & $2.23 \pm 0.68$ & $8.37 \pm 0.59$ \\
$\pi_{\mathrm{cont.}}$-PRL & $0.26 \pm 0.16$ & $0.15 \pm 0.05$ & $1.84 \pm 0.90$ & $2.79 \pm 0.57$ \\
$\pi_{\mathrm{disc.}}$-PRL & $0.36 \pm 0.39$ & $0.21 \pm 0.02$ & $3.05 \pm 0.31$ & $6.29 \pm 0.42$ \\
$\pi$-PRL & $0.57 \pm 0.78$ & $0.18 \pm 0.02$ & $2.35 \pm 1.05$ & $4.43 \pm 1.43$ \\
DiPRL (ours) & $\mathbf{0.07 \pm 0.04}$ & $\mathbf{0.12 \pm 0.01}$ & $\mathbf{0.99 \pm 0.83}$ & $\mathbf{1.76 \pm 1.11}$ \\
\bottomrule
\end{tabular}

    \label{tab:app_continuous_distance}
\end{table}

\clearpage
\section{Domain}
\label{app:domain}

We validate DiPRL and the baselines on both discrete and continuous tasks.

\subsection{Discrete tasks}
\label{sec:discrete-task-descriptions}

We evaluate on five finite-action tasks: three classic-control domains (CartPole-v1, Acrobot-v1, and MountainCar-v0), one MiniGrid navigation domain (DoorKey), and one Box2D control domain (LunarLander-v2).

\subsubsection{CartPole-v1}
The agent controls a cart moving on a one-dimensional track with a pole hinged to the cart. The state is
\begin{equation}
    s_t = (x_t, \dot{x}_t, \theta_t, \dot{\theta}_t),
\end{equation}
where $x_t$ is cart position and $\theta_t$ is pole angle. The action space is $\mathcal{A}=\{\textsc{left},\textsc{right}\}$, corresponding to horizontal forces applied to the cart. The task is to keep the pole upright while the cart remains within the track boundary. The return is the number of balanced time steps, capped at 500, so higher is better. Each step gives a reward for reaching the goal.

\subsubsection{Acrobot-v1}
Acrobot is an underactuated two-link pendulum. Only the joint between the two links is actuated, and the objective is to swing the free end above a target height. The observation contains a trigonometric encoding of both joint angles and their angular velocities,
\begin{equation}
    s_t = (\cos\theta_{1,t},\sin\theta_{1,t},\cos\theta_{2,t},\sin\theta_{2,t},
           \dot{\theta}_{1,t},\dot{\theta}_{2,t}).
\end{equation}
The action space is $\mathcal{A}=\{-1,0,+1\}$, representing negative, zero, or positive torque. The environment gives a per-step penalty until the goal is reached, so better policies have less negative returns.

\subsubsection{MountainCar-v0}
MountainCar controls an underpowered car that must build momentum to reach the goal at the top of a hill. The state is
\begin{equation}
    s_t = (p_t,\dot{p}_t),
\end{equation}
where $p_t$ is position and $\dot{p}_t$ is velocity. The action space is $\mathcal{A}=\{\textsc{left},\textsc{coast},\textsc{right}\}$. The canonical task gives a penalty at each time step until the goal is reached or the episode times out, so higher returns correspond to reaching the goal in fewer steps. During training, distance-based reward shaping is used in the experimental setup; evaluation reports the task return.

\subsubsection{DoorKey}
DoorKey is a MiniGrid navigation task. The agent must navigate a grid, pick up a key, unlock a door, and reach the goal cell. The action space contains discrete navigation and object-interaction actions, including turning, moving forward, picking up the key, and toggling the door. The task has sparse success feedback: unsuccessful episodes receive zero return, while successful episodes receive a positive reward that is larger when the goal is reached earlier.

\subsubsection{LunarLander-v2}
LunarLander-v2 is a discrete-action Box2D landing task. The lander state includes its position, velocity, angle, angular velocity, and leg-contact indicators. The action space is
\begin{equation}
    \mathcal{A}=\{\textsc{noop},\textsc{left engine},\textsc{main engine},\textsc{right engine}\}.
\end{equation}
The objective is to land softly between the landing flags while controlling fuel use and avoiding crashes. Higher episodic return indicates a more stable and efficient landing policy.

\subsection{Continuous tasks}
\label{sec:continuous-task-descriptions}

We also evaluate on four continuous tasks introduced by~\citet{qiu2022programmatic}: HalfCheetah Hurdle, Pusher2D, Ant RandomGoal, and Ant CrossMaze. These tasks use continuous action spaces and require locomotion or manipulation policies that make progress toward task-specific goals. In addition to reward, we report goal distance for these domains, where lower distance indicates better goal reaching.

\subsubsection{HalfCheetah Hurdle}
HalfCheetah Hurdle is a locomotion task in which a half-cheetah robot must move forward while clearing a sequence of hurdles. The observation contains the robot's proprioceptive state together with task-specific information about the next hurdle. The action is a continuous six-dimensional motor command,
\begin{equation}
    a_t \in \mathbb{R}^{6}.
\end{equation}
The reward combines forward progress, jumping behavior, distance to the target position, hurdle-related shaping, a large completion bonus, and collision penalties. The task is solved by reaching the target region beyond the hurdles; higher reward and lower final goal distance indicate better performance.

\subsubsection{Pusher2D}
Pusher2D is a planar manipulation task. A three-joint arm must push a puck to a fixed goal location. The observation includes trigonometric joint encodings, joint velocities, the end-effector position, and the object position. The action is a continuous three-dimensional control vector,
\begin{equation}
    a_t \in \mathbb{R}^{3}.
\end{equation}
The reward is the negative weighted sum of the arm-object distance, the object-goal distance, and a control penalty. Consequently, better policies keep the arm close to the object, move the object toward the goal, and use smoother controls.

\subsubsection{Ant RandomGoal}
Ant RandomGoal is a quadruped locomotion task with a randomly sampled target position in a bounded circular arena. The observation contains the ant's proprioceptive state and the current goal position. The action is a continuous eight-dimensional torque command,
\begin{equation}
    a_t \in \mathbb{R}^{8}.
\end{equation}
The agent receives dense reward for reducing its distance to the sampled goal, with additional healthy-survival reward and control/contact costs. Episodes terminate when the ant reaches the goal or leaves the allowed area. We measure both episodic reward and the final distance to the sampled goal.

\subsubsection{Ant CrossMaze}
Ant CrossMaze uses the same ant morphology but places the agent in a maze-like environment with walls and candidate goal locations. The benchmark variant used in our experiments samples the goal from a small set of maze targets. The observation contains the ant state and the active goal position, and the action is again a continuous eight-dimensional torque command,
\begin{equation}
    a_t \in \mathbb{R}^{8}.
\end{equation}
The task requires both locomotion and navigation: the ant must move through the maze to the active goal while avoiding poor routes and unstable contacts. Reward is based on goal progress, survival, and control/contact penalties; lower final goal distance indicates more reliable navigation.

\section{Declaration of LLM usage}
\label{app:llm}
We used generative AI for proofreading. All scientific content, claims, and results were produced and verified by the authors.

\end{document}